\documentclass[twocolumn,10pt]{asme2ej}

\usepackage{epsfig} 

\title{Runtime Performances Benchmark for Knowledge Graph Embedding Methods}

\author{Angelica Sofia Valeriani
    \affiliation{
	DEIB (Dipartimento di Elettronica, Informazione e Bioingegneria) \\
	Politecnico di Milano\\
	Milan, Italy\\
    Email: angelica.valeriani@mail.polimi.it
    }	
}

\usepackage{url}
\usepackage[hidelinks]{hyperref}
\usepackage{amsmath}
\usepackage{amssymb}

\begin{document}

\maketitle    

\begin{abstract}
{\it
This paper wants to focus on providing a characterization of the runtime performances of state-of-the-art implementations of KGE alghoritms, in terms of memory footprint and execution time. Despite the rapidly growing interest in KGE methods, so far little attention has been devoted to their comparison and evaluation; in particular, previous work mainly focused on performance in terms of accuracy in specific tasks, such as link prediction. To this extent, a framework is proposed for evaluating available KGE implementations against graphs with different properties, with a particular focus on the effectiveness of the adopted optimization strategies. Graphs and models have been trained leveraging different architectures, in order to enlighten features and properties of both models and the architectures they have been trained on. Some results enlightened with experiments in this document are the fact that multithreading is efficient, but benefit deacreases as the number of threads grows in case of CPU. GPU proves to be the best architecture for the given task, even if CPU with some vectorized instructions still behaves well. Finally, RAM utilization for the loading of the graph never changes between different architectures and depends only on the type of graph, not on the model.
}
\end{abstract}



\section{Introduction}

Knowledge Graphs (KGs) are structured representations of real world information. They have found many applications in industry and 
academic settings, and this brought to a considerable research effort towards large-scale information extraction from a variety of 
resources and question answering. A KG is a multi-relational graph composed of entities and relations. In KG, nodes represent entities, 
such as people and places, while edges define specific facts connecting two 
entities with a relation. In fact, each edge is represented as a triple of the form (head entity, relation, tail entity), meaning 
that two entities are connected by a specific relation. KGs can contain millions of entities and billions of facts; some examples can 
be Yago \cite{yago}, FreeBase \cite{freebase} and WikiData \cite{wikidata}. \newline 

Although KGs are effective in representing structured data, the manipulation of triples makes them hard to be managed at a practical level; this issue 
brought to the idea of Knowledge Graph Embedding (KGE), that aims at embedding KG components, including entities and relations into 
continuous vector spaces, in order to simplify the manipulation of data and, at the same time, to preserve the structure of the KG. \newline

Graph embedding learns a mapping from a network to a vector space, while preserving relevant network properties. Graph embeddings are the transformation of property graphs to a vector or a set of vectors. Embedding should capture the graph topology, vertex-to-vertex relationship, and other relevant information about graphs, subgraphs, and vertices. Graphs are a meaningful and understandable representation of data, but there are a few reasons why graph embeddings are needed. First, it must be considered that machine learning on graphs is limited. Second, embeddings are compressed representations, therefore they reduce memory utilization also of very huge graphs, allowing an easier managing.Third, vector operations are simpler and faster than comparable operations on graphs. Vector spaces are more amenable to data science than graphs. Graphs contain edges and nodes, those network relationships can only use a specific subset of mathematics, statistics, and machine learning. On the other hand, vector spaces have a richer toolset from those domains. Additionally, vector operations are often simpler and faster than the equivalent graph operations. \newline

Most of currently available techniques perform embedding task solely on the basis of observed facts. Given a KG, such techniques 
represent at first entities and relations in a continuous vector space, then define a scoring function on each fact to measure its 
plausibility. Embeddings of entities and relations can then be obtained by maximizing the total plausibility of observed facts. During 
this whole procedure, the learnt embeddings are only required to be compatible within each individual fact, and therefore, they cannot 
be predictive enough for downstream tasks. Other techniques use not only facts alone, but also leverage additional information \cite{KnowledgeGraphEmbedding1}. \newline

Real-world knowledge graphs are usually incomplete. For example, Freebase and WikiData are very large but they do not contain all knowledge related to the entities they represent. Knowledge graph completion, or link prediction, is the task that aims to predict new triples \cite{KnowledgeGraphEmbedding3}. This task can be undertaken by using knowledge graph embedding methods, which represent entities and relations as embedding vectors in semantic space, then model the interactions between these embedding vectors to compute matching scores that predict the validity of each triple. Knowledge graph embedding methods are not only used for knowledge graph completion, but the learned embedding vectors of entities and relations are also very useful in a huge amount of emerging ML applications.  
\newline

KGs embedding models are composed of embeddings (i.e., vector representations of entities and relations), as well as interaction mechanism, that models the interaction between 
embedding vectors to compute the matching score of a triple, and predictions, that use the score to predict validity of a triple \cite{KnowledgeGraphEmbedding3}.
Based on the modeling of interaction, a KG embedding model can be classified into one of these three categories: translational-based, neural-network-based, and trilinear-product-based.
Graphs and models have been trained leveraging different architectures, in order to enlighten features and properties of both models and the architectures they have been trained on. The most representative graphs in the state of the art have been selected for this project: \textbf{wn18, wn18rr, fb15k, fb15k-237 and yago3-10}. These graphs differ from the point of view of the number of contained entities (nodes) as well as the number of relations between entities. From the set of considered graphs, wn18 and wn18rr are the less complex graphs, fb15k and fb15k-237 are more complex, while yago3-10 is the subset of one of the greatest graphs today widely used in the state of the art (YAGO) \cite{yago, yago3}. In order to provide a significant overview, the choice was to consider the most representative KGE models in the state of the art. In the context of knowledge graph embedding, there are different categories of models, whose classification is made on the basis  interaction mechanism employed to compute the matching score of a triple; in particular the choice in this project was to consider \textbf{TransE} from the category of translational-based embedding models, \textbf{ConvKB} from the category of neural-network-based embedding models and \textbf{DistMult} from the category of trilinear-product-based embedding models. As already stated, the purpose of the project was to characterize the runtime performance of different KGE methods, w.r.t. the properties of input graphs and the adopted optimization strategies; therefore, the same tests were run in CPU environments, with and without vectorization enabled, and in GPU environment. 
\newline 

For the sake of consistency, the decision was to use the implementation of models taken from AmpliGraph, on GitHub \cite{ampligraph}. The chosen models were all taken considering the most representative models in their category, for example considering features like release time and number of citations. For homogeneity, the decision was to select all the implementations of models from a single framework, in order to guarantee comparability and uniformity, and to minimize the bias introduced by implementation choices. \newline \newline
In Section \ref{overview} of this paper an overview on KGE models, with the explanation of the choice of models is provided. In Section \ref{method} the experimental description of the work and the implementation is described. In section \ref{exp} experimental results are discussed and illustrated, together with a theoretical justification and a critical analysis. Section \ref{end} summarizes the conclusions.

\section{Overview on KGE Models \& Models Choice}
\label{overview}

Knowledge Graphs such as WordNet \cite{wordnet}, Freebase \cite{freebase} and Yago \cite{yago} have been playing a fundamental role in many AI applications, such as relation extraction (RE) or question answering (Q\&A). They usually contain huge amounts of structured data as the form of triples (head entity, relation, tail entity), denoted as \textit{(h,r,t)}, where relation models the relationship between the two entities. In the last few years, many
researchers focused on the knowledge graph completion task, which consists of
predicting relations between entities based on existing triples in a knowledge
graph \cite{KnowledgeGraphEmbedding2}. However, much of the work for knowledge graph completion made in the past decade, based on symbol and logic, is neither tractable nor enough convergent for large scale knowledge graphs. Recently, a powerful approach for this task is to encode every element (entities and relations) of a knowledge graph into a low-dimensional embedding vector space. These methods do reasoning over knowledge graphs through algebraic operations. 
\subsection{Translational-based embedding models} 
Translation-based embeddings aim to ﬁnd vector representations of knowledge graph entities in an embedding space by regarding relations as translation of entities in the space. 
These models translate the head entity embedding by summing with the relation embedding vector, then measuring distance between translated images of head entity and tail entity embedding. 
Even though these translation-based models achieve high performance in knowledge graph completion, they all ignore logical properties of relations. That is, transitive relations and symmetric relations lose their transitivity or symmetricity in the vector space generated by the translation-based models. As a result, the models can not complete only new knowledge with such relations, but also new knowledge with a relation that depends on these relations. 

In most knowledge graphs, transitive or symmetric relations are common. For instance, FB15K, one of the benchmark datasets employed in this work, has a number of transitive and symmetric relations that is about 20\% of its total number of triples. Therefore, the ignorance of logical properties of relations could become a serious problem in knowledge graph completion. The main reason why existing translation-based embeddings cannot reﬂect logical properties of relations is that they do not consider the role of entities. An entity should be a different vector in the embedding space according to its role. Therefore, the solution to preserve the logical properties of relations in the embedding space is to distinguish the role of entities while embedding entities and relations. A role-speciﬁc projection to preserve logical properties of relations in an embedding space can be implemented by projecting a head entity onto an embedding space by a head projection operator and a tail entity by a tail projection operator. As a result, an identical entity is represented as two distinct vectors. 

This idea can be applied to various translation-based models including TransE, TransR, and TransD. According to the previous work \cite{KnowledgeGraphEmbedding2, KnowledgeGraphEmbedding7}, the logical property preserving embeddings achieve the state-of-the-art performance in given task. TransE \cite{transe} is the simplest translation-based graph embedding. It assumes that all vectors of entities and relations lie on a single vector space. Despite its simplicity, TransE shows competitive results in terms of accuracy when compared to other state-of-the-art KGE models \cite{KnowledgeGraphEmbedding1, KnowledgeGraphEmbedding2}. 
\newline \newline
\textbf{TransE} \newline
TransE is the most representative translational distance model \cite{KnowledgeGraphEmbedding2, transe}. It represents both entities and relations as vectors in the same space. TransE enforces the concept of using a geometric interpretation of the latent space, by requiring that the tail embeddings lie close to the sum of the head and relation embeddings, according to the chosen distance function. 
It means, given a fact\textit{(h, r, t)}, the relation is interpreted as a translation vector r so that the embedded entities h and t, head and tail, can be connected via r with low error if the condition 
\textbf{h + r $\approx$ t} is verified. The scoring function is defined as the negative distance between h+r and t, and is expected to be large if \textit{(h, r, t)} holds. Due to the nature of translation, TransE is not able to correctly handle one-to-many and many-to-one relations, as well as symmetric and transitive relations. This issue comes to the point that, considering considering for example the one-to-many relation with the head \textit{h} fixed, then all the tails \textit{t} in that relation with the head \textit{h} will have a very similar representation, even if they can be totally different entities. The solution to this problem is to allow the entities to play different roles according to relations, but TransE is unable to do it. Despite that, according to experimental results of previous works, TransE results competitive in terms of accuracy when compared to other KGE models, while obtaining outstanding performances in terms of time convergence \cite{KnowledgeGraphEmbedding1, KnowledgeGraphEmbedding2}.


\subsection{Neural-network-based embedding models}
Neural-network-based embedding models use a nonlinear neural network as universal approximators to compute the matching score for a triple.
Neural Networks learn parameters such as weights and biases, that they combine with the input data in order to recognize significant patterns \cite{KnowledgeGraphEmbedding2}. 
Deep neural networks usually organize parameters into separate layers, generally interspersed with non-linear activation functions.
In particular, convolutional neural networks use one or multiple convolutional layers: each of these layers performs convolution on the input data (e.g. the embeddings of the KG elements in a training fact) applying low-dimensional filters. 
The result is a feature map that is usually then passed to additional dense layers in order to compute the fact score. 
ConvKB \cite{convkb} is the most representative model in this category, with a high number of references in the literature, and it outperforms previous state-of-the-art models several benchmark link prediction datasets \cite{KnowledgeGraphEmbedding1, KnowledgeGraphEmbedding3}. 
\newline \newline
\textbf{ConvKB} \newline
ConvKB is a representative convolutional model, that represents entities and relations as same-sized one-dimensional embeddings \cite{KnowledgeGraphEmbedding2, convkb}. It 
employees a convolutional neural network, to capture global relationships and transitional characteristics between entities and relations 
in knowledge basis. Each triple \textit{(h, r, t)} is represented as a three-column matrix, in which each column vector represents a triple 
element. It concatenates all embeddings of a fact \textit{(h, r, t)} into a matrix [\textbf{h}; \textbf{r}; 
This input is forwarded to a convolutional layer 
with multiple filters, resulting in an output feature map. Feature maps are concatenated into a single feature vector representing the 
input triple. The feature map is let through a dense layer with weights, resulting in the fact score. The score is then used to predict 
whether the triple is valid or not. This architecture can be seen as a binary classifier. Formally, the definition of ConvKB score function \(\mathnormal{f}\) as follows \[\mathnormal{f}(\textit{h,r,t}) = concat(\textit{g} ([\textbf{\textit{v}}_{\textit{h}},\textbf{\textit{v}}_{\textit{r}},\textbf{\textit{v}}_{\textit{t}}] \ast \textbf{$\Omega$})) \cdot \textbf{w}\] where \textbf{$\Omega$} and \textbf{w} are shared parameters, independent of \textit{h, r and t}; $\ast$ denotes a convolution operator; and \textit{concat} denotes a concatenation operator. Experiments show that ConvKB achieves better link prediction performance than previous state-of-the-art embedding models on several benchmark datasets, such as WN18RR and FB15K-237 \cite{KnowledgeGraphEmbedding1, KnowledgeGraphEmbedding3, convkb}.


\subsection{Trilinear-product-based embedding models}
Trilinear-product-based embedding models compute scores by using trilinear product between head, tail and relation embeddings. These models solve a tensor decomposition problem with the matching score of each triple modeled as the result of a trilinear product, i.e., a multilinear map with three variables corresponding to the embedding vectors \textbf{h}, \textbf{t}, and \textbf{r} of head entity \textit{h}, tail entity \textit{t}, and relation \textit{r}, respectively  \cite{KnowledgeGraphEmbedding2, KnowledgeGraphEmbedding3}. The trilinear-product-based score function for the three embedding vectors is denoted as $\langle$ \textbf{h, t, r} $\rangle$. DistMult \cite{distmult} is the simplest model in this category. It embeds each entity and relation as a single real valued vector, but it is still able to obtain competitive results in the link prediction task, thus it is vastly employed in literature \cite{KnowledgeGraphEmbedding2, KnowledgeGraphEmbedding9}.  
\newline \newline
\textbf{DistMult} \newline
DistMult is the simplest model in the category of trilinear-product-based models and a semantic matching model \cite{KnowledgeGraphEmbedding2, distmult, KnowledgeGraphEmbedding9}.
The main idea is to represent the entities and relations in a vector space, and to use machine learning technique to learn the continuous representation of the knowledge graph in the latent space. RESCAL \cite{KnowledgeGraphEmbedding9} is one of the earliest studies on matrix factorization based knowledge graph embedding models, using a bilinear form as score function, in the field of trilinear-product-based models. DistMult simplifies RESCAL by only using a diagonal matrix. It embeds each entity and relation as a single real-valued vector. It exploits similarity-based scoring functions and forces all relation embeddings to be 
diagonal matrices, consistently reducing the space of parameters to be learnt, resulting in an easier model to train. The scoring function becomes commutative, which amounts to treat all relations as symmetric. 
For each relation \textit{r}, it introduces a vector embedding \(r \in \mathbb{R}^{\textit{d}}\) and requires \(\textbf{M}_{\textit{r}} = diag(\textbf{r})\). The scoring function is hence defined as \[\mathnormal{f}_{\textit{r}}(\textit{h,t}) = \textbf{h}^{\top} diag(\textbf{r}) \textbf{t} =  \sum_{i=0}^{\textit{d}-1} [\textbf{r}]_{\textit{i}} \cdot [\textbf{h}]_{\textit{i}} \cdot [\textbf{t}]_{\textit{i}} \] This score captures pairwise interactions between only the components of \textit{h} and \textit{t} along the same dimension \cite{KnowledgeGraphEmbedding1}. Moreover its score function is symmetric, with the same scores for triples \textit{(h, r, t)} and \textit{(t, r, h)}. Therefore, it cannot model asymmetric data for which only one direction is valid. However, thanks to its simplicity and its competitive link prediction performance, DistMult is most employed  trilinear-product-based model in literature \cite{KnowledgeGraphEmbedding2, distmult, KnowledgeGraphEmbedding9}.

\section{Methodology}
\label{method}

As anticipated in Section \ref{overview}, the purpose of this project was to characterize the runtime performances of different KGE methods, w.r.t. the properties of input graphs and the adopted optimization strategies. 
To this aim, a python framework has been implemented to easily handle different KGE experiments, supporting multiple graphs, models, and configuration settings.
Such framework has been built on top of AmpliGraph \cite{ampligraph}, an open source library for supervised learning on knowledge graphs.
AmpliGraph is, in turn, based on TensorFlow \cite{tensorflow}, a state-of-the-art end-to-end open source platform for machine learning, widely employed both in industry and academia.
The proposed python framework allows users to configure the execution of KGE experiments by setting a series of parameters. 

First, the user can select the KGE model to be employed in the experiment choosing among the three models presented in Section \ref{overview}: TransE, DistMult and ConvKB.
Then, the input graph can be selected among the following knowledge graphs, widely employed in literature: WN18, WN18RR, FB15K, FB15K-237 and YAGO3-10.
WN18 is a subset of WordNet \cite{wordnet} which consists of 18 relations and 40,943 entities. Most of the 151,442 triples consist of hyponym and hypernym relations and, for such a reason, WN18 tends to follow a strictly hierarchical structure. FB15K is a subset of Freebase \cite{freebase} which contains about 14,951 entities with 1,345 different relations. A large fraction of content in this knowledge graph describes facts about movies, actors, awards, sports, and sport teams. WN18RR and FB15K-237 are, respectively, subsets of WN18 and FB15K deprived of "inverse relations". 
Two relations \textit{r1} and \textit{r2} are "inverse relations" if the existence of the triple \textit{(h, r1, t)} implies the existence of the triple \textit{(t, r2, h)}.
Finally, YAGO3-10 is a subset of YAGO3 \cite{yago3} which consists of entities with a minimum of 10 relations each. It has 123,182 entities and 37 relations. Most of triples deal with descriptive attributes of people, such as citizenship, gender, and profession.
Table \ref{KGproperties} presents the features of the selected graphs, in terms of number of entities, number of relations, and size of train, validation and test sets. 
Train and validation sets are employed when training KGE models, while test set is employed to measure inference performances of the trained model.
In this project, both graphs and implementations of models have been taken from AmpliGraph, to guarantee consistency and comparability, trying to minimize bias introduced by implementation choices. 

\begin{table}[h!]
  \caption{Knowledge Graph Properties}
  \label{KGproperties}
  \center
  \resizebox{0.4\textwidth}{!}{%
  \begin{tabular}{l|l|l|l|l|l}
    \hline
    \textbf{Graph} & \textbf{Entities} & \textbf{Relations} & \textbf{Train} & \textbf{Validation} & \textbf{Test}\\
    \hline
    WN18	& 40943	& 18 & 141442 & 5000 & 5000 \\
    WN18RR	& 40943	& 11 & 86835 & 3034 & 3134 \\
    FB15K	& 14951	& 1345 & 483142 & 50000 & 59071 \\
    FB15K-237	& 14541	& 237 & 272115 & 17535 & 20466 \\
    YAGO3-10	& 123182	& 37 & 1079040 & 5000 & 5000 \\
  \end{tabular}}
\end{table}

After selecting the model and the graph, the proposed framework allows to set the number of threads employed the experiment.
In particular, the number of threads indicated by the user is used to set the TensorFlow option \textit{intra\_op\_parallelism\_threads}, which defines the number of threads used to parallelize the execution of individual operations scheduled in the TensorFlow Graph \cite{proto}. 
On the other hand, the option \textit{inter\_op\_parallelism\_threads} defines the size of the pool of threads different operations are enqueued on.
We set $\text{\textit{inter\_op\_parallelism\_threads}}= 2$ for each experiment, according to reccomendations by Intel on how to maximize TensorFlow performance \cite{intel2}. 

The proposed framework also allows to choose whether to monitor the RAM usage during the experiment, or not.
If the RAM usage is monitored, the peak RAM usage is saved both immediately after the load of the graph and at the end of the experiments. 
In this way, it is possible to calculate the RAM usage of the selected model for a given graph, without considering the RAM employed to store such graph.
The framework also accepts as input the number of epochs (i.e., iterations of the training loop) to be used for training the KGE model.
Finally, the user can provide a seed number for random numbers generator, in order to guarantee replicability of the experiments.
Other settings, such as the number of batches, the loss function, or the optimizer used to minimize it during training, can be easily set by modifying the "setup" portion at the beginning of the source code.

For each experiment, the framework measures the training time, and inference times for entities, relations, and scores of triples composing the test set. 
For each task, the framework measures both the wallclock time and the CPU time; in particular, we have to consider that the wall clock time is the actual amount of time taken to perform a job, while CPU time measures the total amount of time the CPU spent running the code or anything requested by the code.
The results of each experiment are saved as a row in a .csv file, which can be later employed to create plots and extract insight from the results. 
All the procedures were automatized both for the execution of the experiments and for the realization of plots.
All plots presented in the document and on the GitHub repo \cite{repo} have been realized through Pandas \cite{pandas} and Matplotlib \cite{matplotlib}, by selecting the most significant results and studying the best way to present them.

\subsection{Experimental design and setup}
During this project, multiple experiments were conducted employing each possible combination of KGE method and input graph.
Moreover, for each of such combinations, three experiments were run employing three different numbers of threads.
Each experiment has been done twice, both monitoring and not monitoring the RAM usage, in order to avoid bias on the measurement of execution times.
The same experiments were run in CPU environments, with and without vectorization enabled, and in a GPU environment.
Thus, the total number of experiments (excluding preliminary tests) is 270, totalizing 630+ hours of execution time.

In addition to the parameters provided by the user, other options can be set to configure the execution of experiments.
The decision was to determine a suitable setting for all the KGE models and all the graphs; after a number of preliminary tests, and time passed studying this topic, the common setting was determined. 
The choice of selecting a common setting derives from the need of getting an efficient and consistent way to compare the performances results obtained by different models (trained on different graphs), across the investigated  architectures. 
This is effective to show how different kinds of parallelization influence runtime performances of KGE experiments. 
In particular, the chosen settings are:
\begin{itemize}
  \item \textit{Embedding space dimensionality: \textbf{256}};
  \item \textit{Number of iterations of the training loop (number of epochs): \textbf{500}};
  \item \textit{Number of negatives generated at runtime during training for each positive (ETA): \textbf{2}};
  \item \textit{Number of batches in which the training set must be split during the training loop: \textbf{100}};
  \item \textit{Loss function to use during training: \textbf{pairwise, max-margin loss}};
  \item \textit{Optimizer used to minimize the loss function: \textbf{`Adam'}} \cite{adam};
  \item \textit{Learning rate of the optimizer: \textbf{0.01}}.
\end{itemize}
From preliminary tests, we noticed that the choice of the optimizer greatly impact the runtime performances of the explored models.
We chose to employ the 'Adam' optimizer \cite{adam}, which is computationally efficient, has little memory requirements, is well suited for problems that are large in terms of data and/or parameters, and is vastly employed in previous work on KGE \cite{KnowledgeGraphEmbedding1, KnowledgeGraphEmbedding2, KnowledgeGraphEmbedding4}.

Baseline experiments on CPU (without enabling vectorization) were conducted on a server equipped with two Intel Xeon E5-2680 v2 @2.8GHz (for a total of 20 cores and 40 threads) and 378GB of RAM. For such tests we employed a number of threads equal to 8, 16 and 32.
All the other tests (leveraging vectorized instructions or GPU) were run on a server equipped with an Intel Core i7-6700 @3.4GHz (4 cores, 8 threads), 32GB of RAM, and a Nvidia GeForce GTX 960 (1024 cores, 1126 MHz, 2 GB of GDDR5 @ 1750 MHz).
The choice of running experiments leveraging vectorized instructions on the smaller server is tied to the fact that the Intel Core i7-6700 supports SSE4.1, SSE4.2, AVX, AVX2 and FMA instructions, while the Intel Xeon E5-2680 v2 only supports SSE4.1, SSE4.2 and AVX instructions, thus the impact of enabling vectorization on the runtime performance would have been less evident.
Table \ref{tools} lists the specific versions of the tools employed for the execution of tests and the extraction of results.

\begin{table}[h!]
  \center
  \caption{Employed Tools}
  \label{tools}
  \resizebox{0.2\textwidth}{!}{%
  \begin{tabular}{l|l}
    \hline
    \textbf{Tool} & \textbf{Version}\\
    \hline
    python	& 3.7.7	\\
    ampligraph	& 1.3.1	\\
    tensorflow	& 1.15.0 \\
    Keras-Applications	& 1.0.8	\\
    Keras-Preprocessing	& 1.1.0	\\
    CUDA Toolkit & 10.0.130 \\
    cuDNN & 7.6.5 \\
    numpy	& 1.18.1\\
    pandas	& 1.0.3 \\
  \end{tabular}}
\end{table}
\section{Experimental results}
\label{exp}
In this section experimental results obtained with every tested architecture are presented and analyzed. In particular, Subsection \ref{exp1} describes and comments the results obtained by test performing in a standard CPU environment, without vectorization enabled. Subsection \ref{exp2} focuses on the improved environment, that is CPU environment with some vectorized instructions enables. Subsection \ref{exp3} describes instead same tests trained and analyzed in the GPU environment. Finally, Subsection \ref{exp4} makes a critical and complete analysis of all the tested architectures, commenting their features and obtained results.
\subsection{CPU architecture without vectorization}
\label{exp1}
\begin{figure} 
\centerline{\psfig{figure=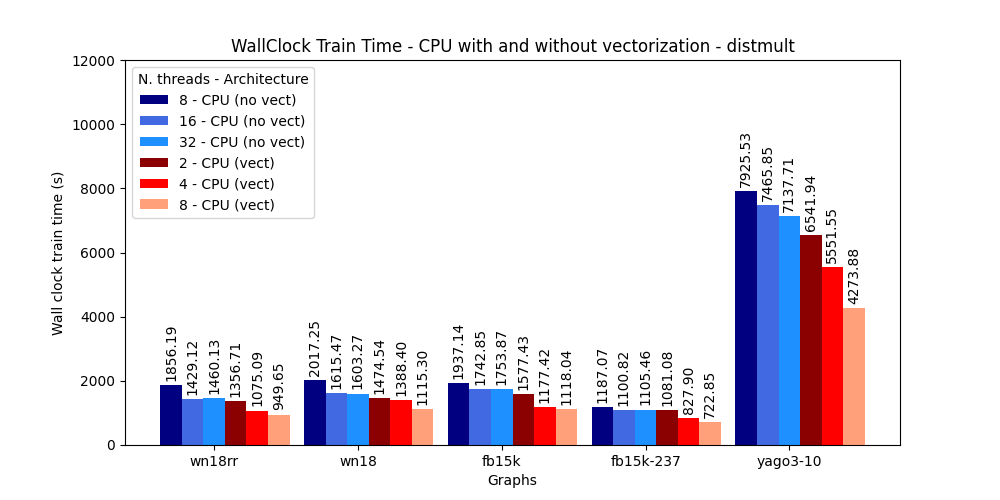,width=3.34in}}
\caption{DistMult: comparison between CPU architecture without vectorization and with vectorization considering wall clock train time (in seconds) on y-axis.}
\label{fig_example1.ps}
\end{figure}

Considering the standard CPU architecture, tests have been performed with a number of threads of 8, 16 and 32, in case of the current architecture. In general, it can be noticed that increasing the number of threads wall clock train time tends to decrease, while CPU train always increases. With models TransE and DistMult, the variation between the measurements of wall clock train time is significant considering the difference between 8 and 16 threads used for training. See figure~\ref{fig_example1.ps} and figure~\ref{fig_example3.ps} Considering TransE illustrations, the difference in structure of the graphs is evident in the plots of CPU time and wall clock time, in particular considering that, with each model tested, the CPU time is always at least twice the one of other graphs, and can become also of the order of five/six times greater, and the same happens for the wall clock time measured values. This is due to the fact that yago3-10 has a number of entities that is of the order of 10 times greater with respect to fb15k and fb15k-237, while it is of the order of 3 times greater with respect to wn18 and wn18rr. Each of the entities has minimum ten relations, therefore it is a very consistent graph compared to the others. Every graph has been trained with all the models, TransE, ConvKB and DistMult, having the fixed settings described in Section \ref{method}. Nevertheless, passing from 16 threads to 32 threads, with TransE and DistMult, this variation in wall clock train time becomes negligible. Model ConvKB keeps a regular behavior, considering the described trends; wall clock train time tends to decrease, while CPU train time always tends to increase. Wall clock and CPU times measured for inference of relations and entities are infinitesimal values, therefore the difference, changing the number of threads is negligible. \newline

The most significant variation is the one concerning the time of training. These results are supported by the theory; in fact, considering that wall clock time is the actual amount of time taken to perform a job. This is equivalent to timing your job with a stopwatch and the measured time to complete the task can be affected by anything else that the system happens to be doing at the time. On the contrary, CPU time measures the total amount of time the CPU spent running the code or anything requested by the code. This includes kernel time. The job requires less time to be executed, as more threads are running in parallel, nevertheless CPU stays busy for a huger amount of time, considering the time required for the parallelization between threads. In the GitHub repo \cite{repo}, all plots can be found, representing both CPU time and wall clock time for one model and all the tested graphs, and the same for one graph and all the tested models. \newline

Considering the datasets structure, some considerations can be made. Notice that all these considerations regard the behavior of the models with the tested graphs in the moment in which convergence is reached, and inference in subsequent steps can be performed. The illustrated plots don’t represent CPU time and wall clock time in which the graph reaches convergence in the given model. As we already said, settings were kept common in order to guarantee uniformity in the measurement of performance with respect to the architecture used. \newline 

The structure of the model has a great influence on the behavior on the graph. In particular, considering that deeper models are more difﬁcult to optimize, we have to consider that for datasets with low average relation speciﬁc in degree (like WN18RR and WN18), a shallow model like DistMult might sufﬁce for accurately representing the structure of the network. On the contrary, deeper models such as ConvKB have an advantage to model more complex graphs (e.g. FB15k and FB15k-237), but that shallow models such as DistMult have an advantage to model less complex graphs (e.g. WN18 WN18RR). As noted by Toutanova and Chen (2015), WN18 and FB15k are easy because they contain many reversible relations \cite{KnowledgeGraphEmbedding8}. Nevertheless, WN18RR and FB15k237 are created not to suffer from this reversible relation problem in WN18 and FB15k, for which the knowledge base completion task is more realistic. ConvKB applies the convolutional neural network to explore the global relationships among same dimensional entries of the entity and relation embeddings, so that ConvKB generalizes the transitional characteristics in the transition based embedding models. Experimental results show that our model ConvKB outperforms other state-of-the-art models on two benchmark datasets WN18RR and FB15k-237 \cite{convkb}. The simple models TransE outperforms other approaches on WN18 in terms of the Mean metric. This may be because the number of relations in WN18 is quite small so that it is acceptable to ignore the different types of relations. Overall, TransE is the runner upon FB15k. However, its relative superiorities on one-to-many and many-to one relations are not as good as those on one-to-one relations. In WN18 and WN18RR we observe a rather different situation. This time, symmetric relations are easily handled by most models, with the notable exceptions of TransE and ConvKB \cite{transe}. On WN18RR, the good results on symmetric relations balance, for most models, sub-par performance on anti-symmetric relations. In YAGO3-10 once again TransE and ConvKB having a hard time handling symmetric relations; on these relations, most models actually tend to behave a little worse \cite{yago3}.
\subsection{CPU architecture with vectorization}
\label{exp2}
Same tests performed on CPU architecture without vectorization have been performed on CPU architecture with vectorization enabled, with a number of threads of 2, 4 and 8, due to architectural constraints.
Note that TensorFlow uses all the threads available in the machine; even if setting the number of threads, the tool uses all the machine cores trying to keep on the whole a behavior in line with the number of set threads. For example, this can be noticed in the case of 4 threads with yago3-10 in the case of CPU train time measured with TransE in this architecture. CPU train time is greater in this case only due to overhead; in fact, the wall clock train time behaves consistently with the described trend (i.e. superior to the time measured with 8 threads, inferior to the one measured with 2 threads). \newline

In general, it can be noticed also in this environment that the wall clock train time always tends to increase, while the CPU train time always decreases by increasing the number of threads. Peculiarities can regard the single model trend; for example, the CPU time measured in case of model ConvKB shows a very small delta in the times measurements despite the variation of the number of threads. See figure~\ref{fig_example6.ps} and figure~\ref{fig_example12.ps}. On the contrary, for the wall clock train time in all the models there is a significant delta in the decreasing of the measured value, by increasing the number of threads. Same considerations about the models, considering the datasets’ structure and properties can be made. \newline

Even if trends in CPU time and wall clock train time are the same in comparison to the measured values in CPU environment without vectorization, it can be noticed that in this architecture the time values are always inferior with respect to the other times. CPU architecture with vectorization enabled is faster due to parallelism performed at level of data. Performing a task, in this environment a single instruction works on vectors of data and the parallelism is more and more efficient on performances as the inner factors allow to do so (i.e. length of operand vectors, structural hazards among operations, data dependences). Therefore, in general, all the measured times in CPU architecture with vectorization enabled are inferior with respect to the basic CPU architecture.
\begin{figure} 
\centerline{\psfig{figure=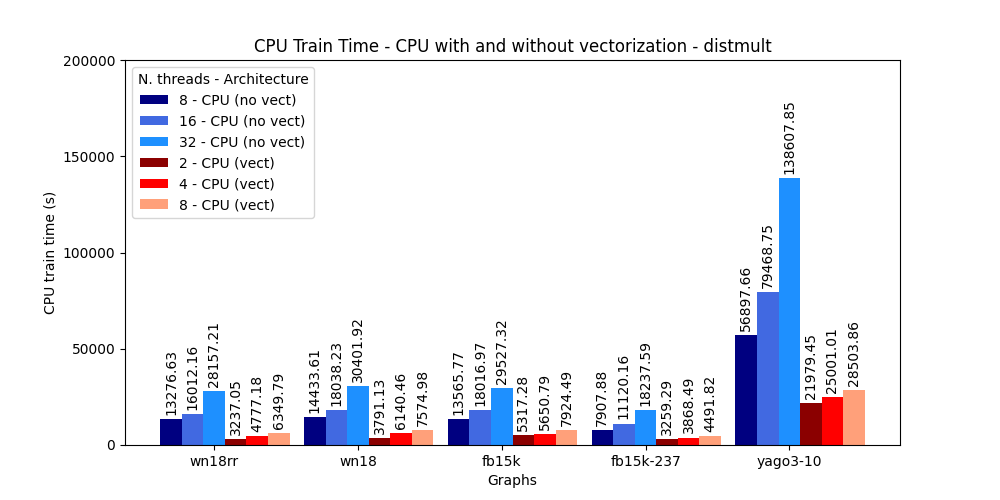,width=3.34in}}
\caption{DistMult: comparison between CPU architecture without vectorization and with vectorization considering CPU train time (in seconds) on y-axis.}
\label{fig_example2.ps}
\end{figure}
\begin{figure} 
\centerline{\psfig{figure=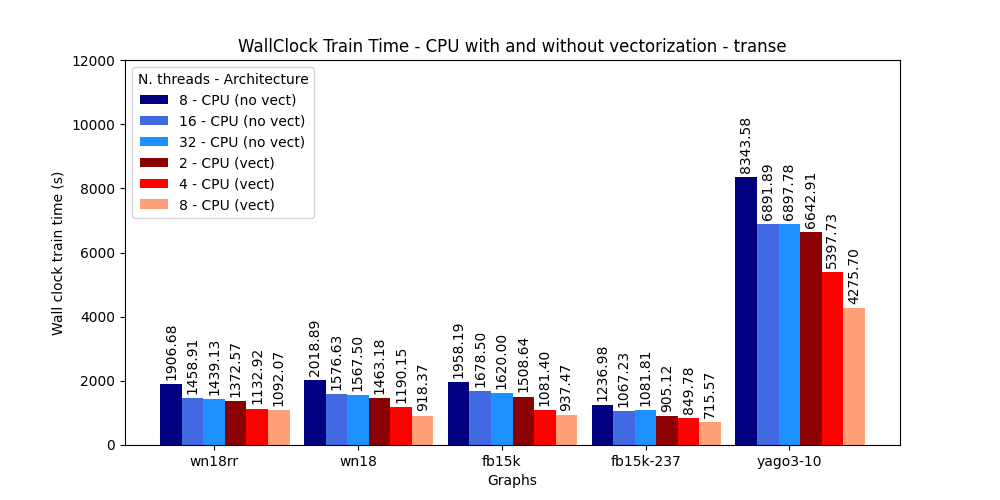,width=3.34in}}
\caption{TransE: comparison between CPU architecture without vectorization and with vectorization considering wall clock train time (in seconds) on y-axis.}
\label{fig_example3.ps}
\end{figure}
\begin{figure} 
\centerline{\psfig{figure=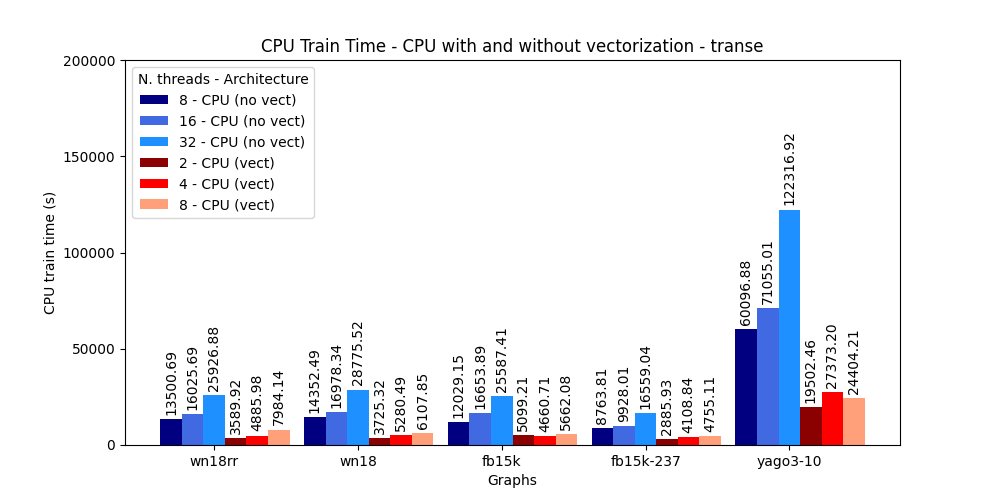,width=3.34in}}
\caption{TransE: comparison between CPU architecture without vectorization and with vectorization considering CPU train time (in seconds) on y-axis.}
\label{fig_example4.ps}
\end{figure}
\begin{figure} 
\centerline{\psfig{figure=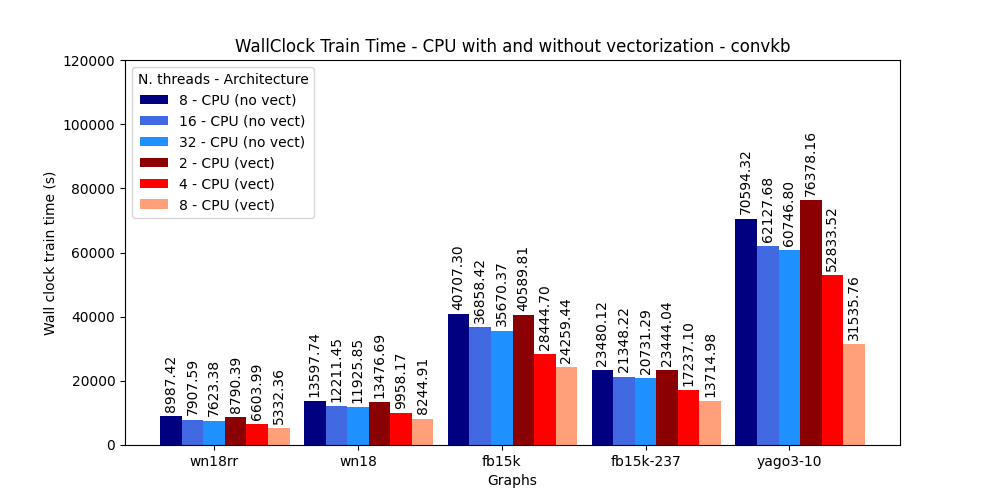,width=3.34in}}
\caption{ConvKB: comparison between CPU architecture without vectorization and with vectorization considering wall clock train time (in seconds) on y-axis.}
\label{fig_example5.ps}
\end{figure}
\begin{figure} 
\centerline{\psfig{figure=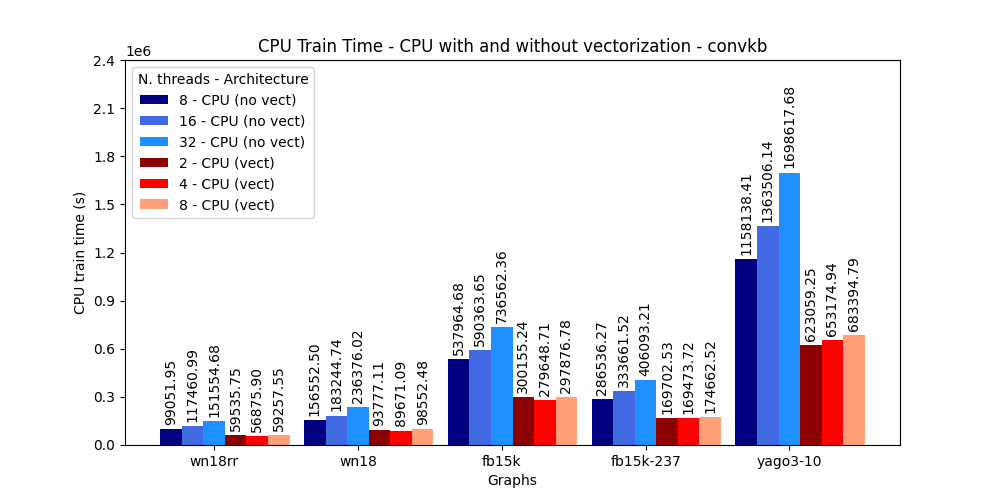,width=3.34in}}
\caption{ConvKB: comparison between CPU architecture without vectorization and with vectorization considering CPU train time (in seconds) on y-axis.}
\label{fig_example6.ps}
\end{figure}
\subsection{GPU architecture}
\label{exp3}
Same tests previously described have been taken in a GPU environment. The selected graphs, wn18, wn18rr, fb15k, fb15k-237 and yago3-10, have been trained with the most representative models of the main categories of models in the field of knowledge graph embedding; TransE, ConvKB and DistMult. \newline

In this environment, a great difference in performances can be noticed, that is, for every model, both wall clock train time, CPU train time and GPU train time remain constant, even changing the number of threads. In this environment, for architectural constraints, the graphs were trained with 2, 4 and 8 threads. Even considering threads variation, wall clock train time, CPU train time and GPU train time don’t change, because of the way in which TensorFlow internally manages cores’ internal division. When the number of threads is set, we are determining the number of resources given for the parallel execution of single tasks, and this is exactly what is done with GPU. \newline

When the training is performed, in case we are on CPU only, the check on the effective number of threads needed is made and processes are spread along the given number of threads. In case of GPU, instead, in phase of training, the greater amount of threads that is allowed is not significant and is not influent. On the whole we have to consider that GPU is faster than CPU, therefore the CPU can be considered a bottleneck. All the measured times are inferior with respect to the same execution performed in a CPU environment. 
\begin{figure} 
\centerline{\psfig{figure=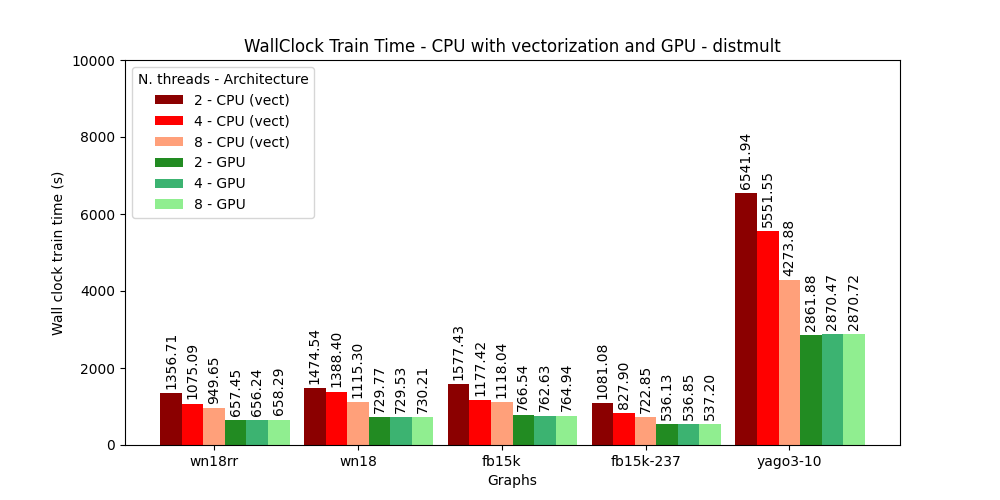,width=3.34in}}
\caption{DistMult: comparison between CPU architecture with vectorization and GPU architecture considering wall clock train time (in seconds) on y-axis.}
\label{fig_example7.ps}
\end{figure}
\begin{figure} 
\centerline{\psfig{figure=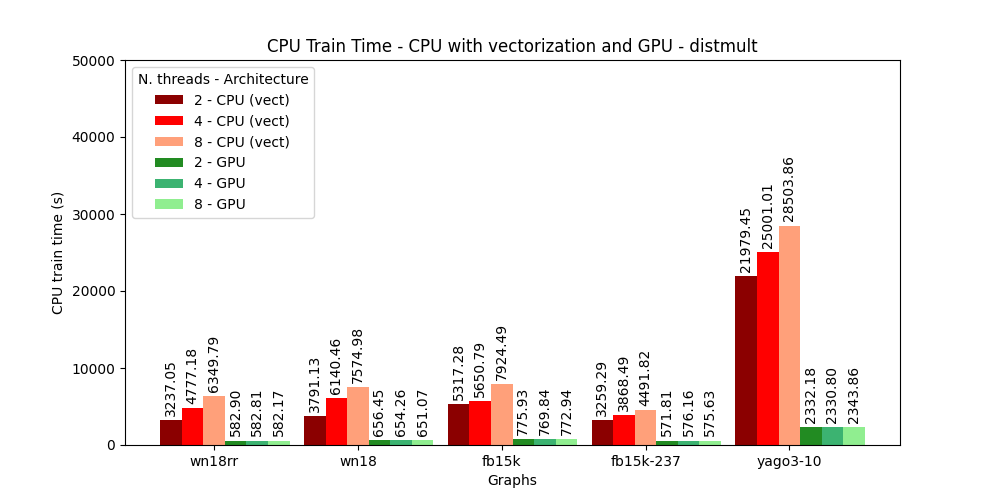,width=3.34in}}
\caption{DistMult: comparison between CPU architecture with vectorization and GPU architecture considering CPU train time (in seconds) on y-axis.}
\label{fig_example8.ps}
\end{figure}
\begin{figure} 
\centerline{\psfig{figure=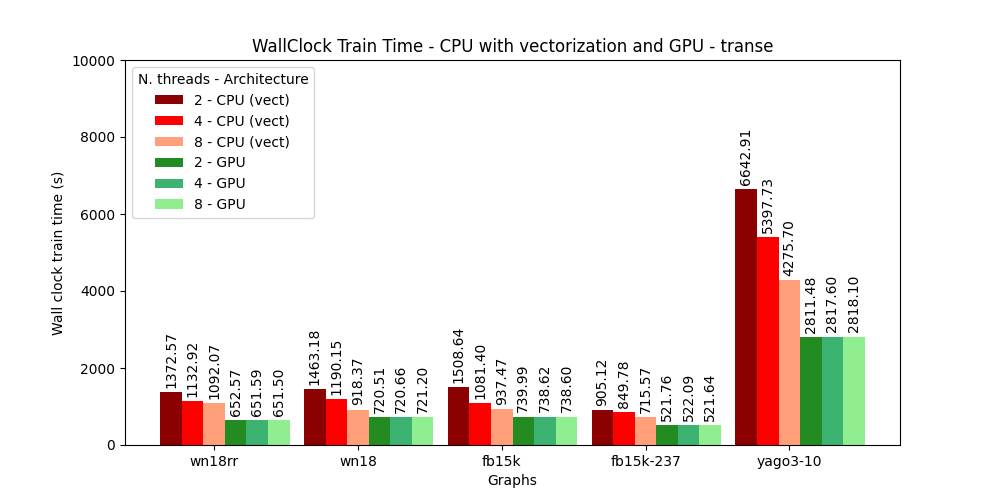,width=3.34in}}
\caption{TransE: comparison between CPU architecture with vectorization and GPU architecture considering wall clock train time (in seconds) on y-axis.}
\label{fig_example9.ps}
\end{figure}
\begin{figure} 
\centerline{\psfig{figure=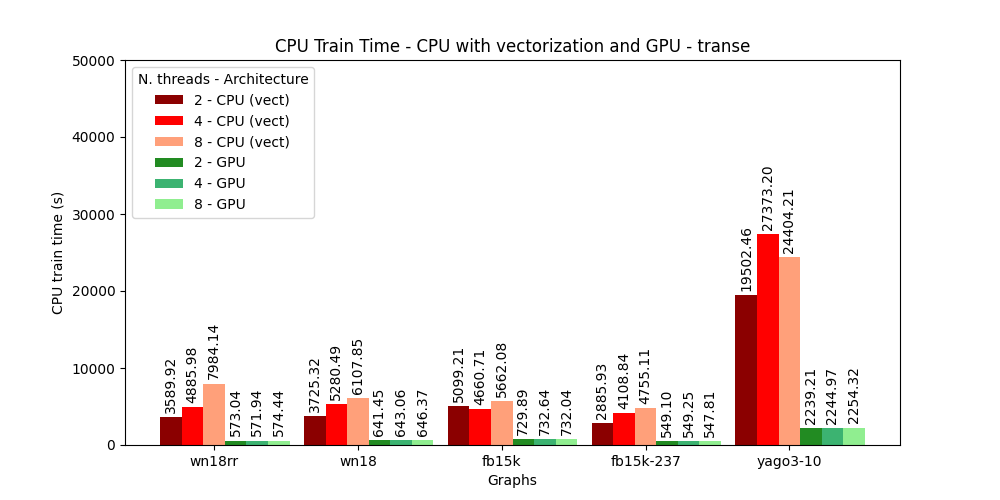,width=3.34in}}
\caption{TransE: comparison between CPU architecture with vectorization and GPU architecture considering CPU train time (in seconds) on y-axis.}
\label{fig_example10.ps}
\end{figure}
\begin{figure} 
\centerline{\psfig{figure=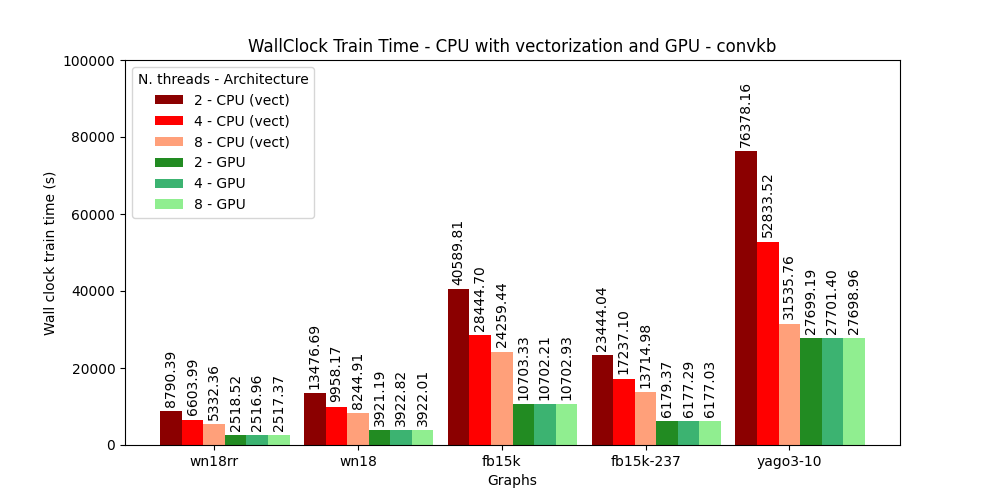,width=3.34in}}
\caption{ConvKB: comparison between CPU architecture with vectorization and GPU architecture considering wall clock train time (in seconds) on y-axis.}
\label{fig_example11.ps}
\end{figure}
\begin{figure} 
\centerline{\psfig{figure=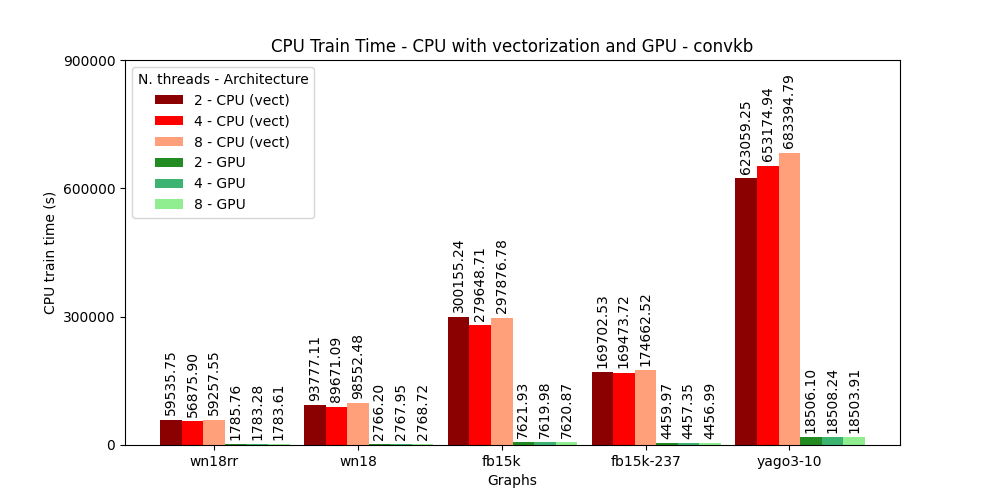,width=3.34in}}
\caption{ConvKB: comparison between CPU architecture with vectorization and GPU architecture considering CPU train time (in seconds) on y-axis.}
\label{fig_example12.ps}
\end{figure}
\subsection{Comparison}
\label{exp4}
\begin{figure} 
\centerline{\psfig{figure=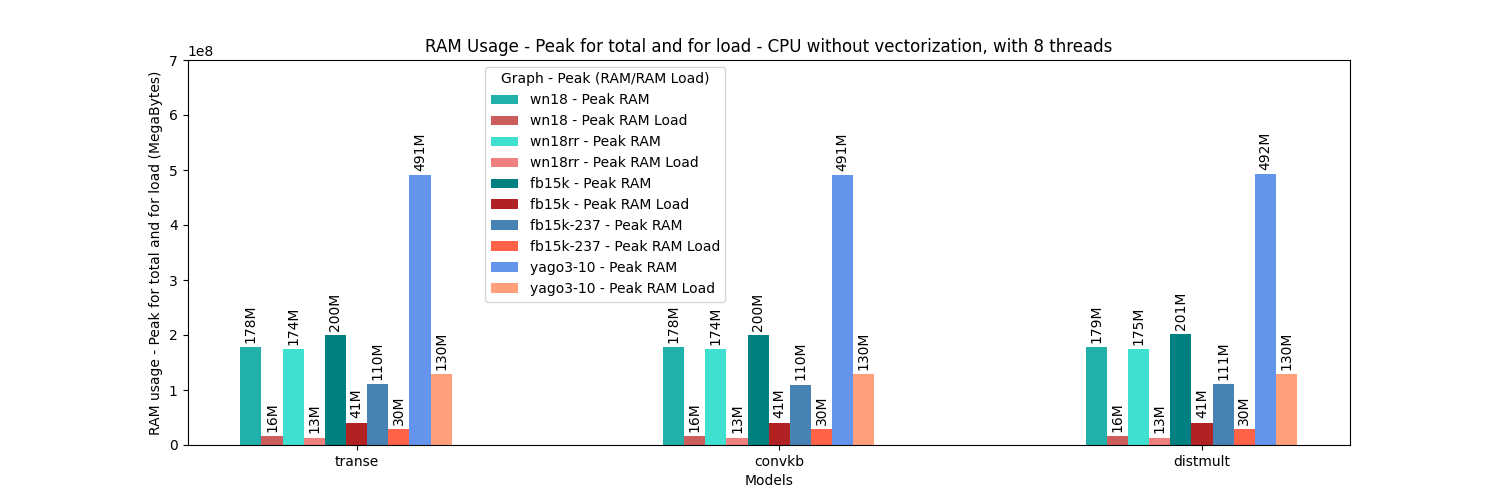,width=3.34in}}
\caption{On the y-axis: RAM usage (in MegaBytes) for total peak and peak of loading graph. On the x-axis: all the tested graphs, considering a number 8 threads, with all the models. Architecture is CPU without vectorization.}
\label{fig1.ps}
\end{figure}

\begin{figure} 
\centerline{\psfig{figure=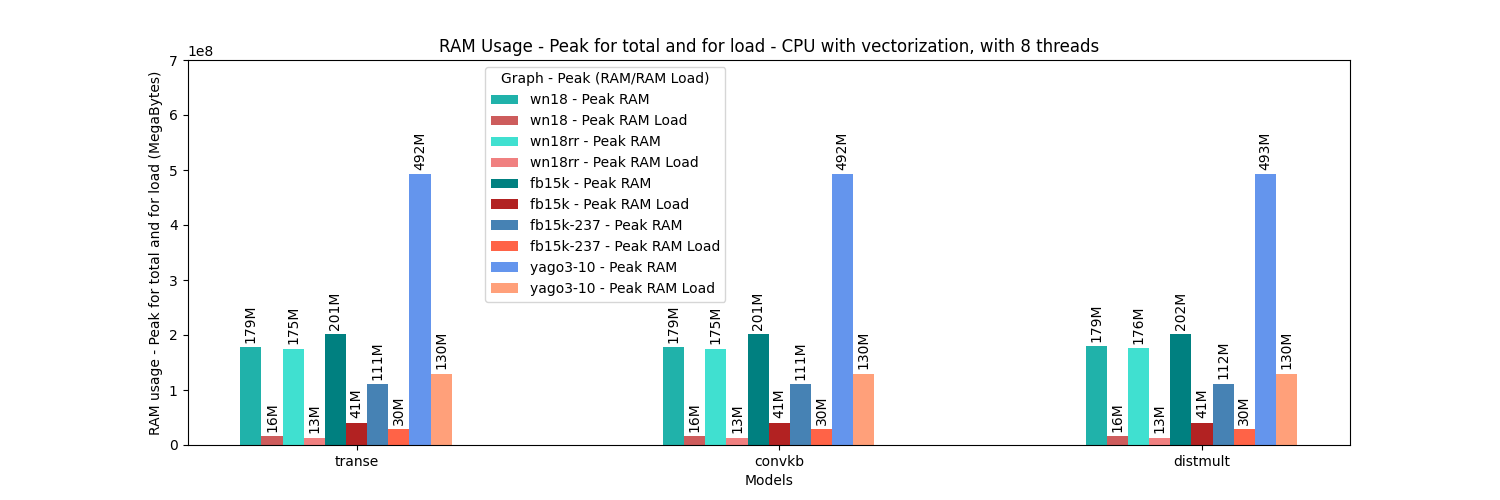,width=3.34in}}
\caption{On the y-axis: RAM usage (in MegaBytes) for total peak and peak of loading graph. On the x-axis: all the tested graphs, considering a number 8 threads, with all the models. Architecture is CPU with vectorization.}
\label{fig2.ps}
\end{figure}

It is evident in all figures in this Section, how the CPU train time tends to increase considering an architecture with only a CPU, while it is constant in case of a GPU architecture. First of all, considering a CPU environment, it can be noticed that vectorization reduced the measured times, both for wall clock train time and CPU train time. Peculiar behaviors of the different models are related to the characteristics and properties of graphs, but on the whole, it can be stated that CPU architecture with vectorization enabled is faster with respect to CPU architecture without vectorization. This is due to the intrinsic parallelism of data that is performed in a vectorized environment. See figures in this section, till figure~\ref{fig_example6.ps}.
Considering RAM usage, it can be noticed that the peak for both the total process, and the load of the graph (i.e. yago3-10 is always the graph with the greatest peak RAM of loading because of its density and dimension) are both very similar in the two environments. \newline CPU architecture with vectorization shows to have a similar RAM usage with respect to the environment without vectorization, as a result of the experiment in this document. In the plots, the total peak of RAM is shown in the blue shadows, while the peak of loading graph is shown in the red shadows. See figure~\ref{fig1.ps} and figure~\ref{fig2.ps}.  \newline

\begin{figure} 
\centerline{\psfig{figure=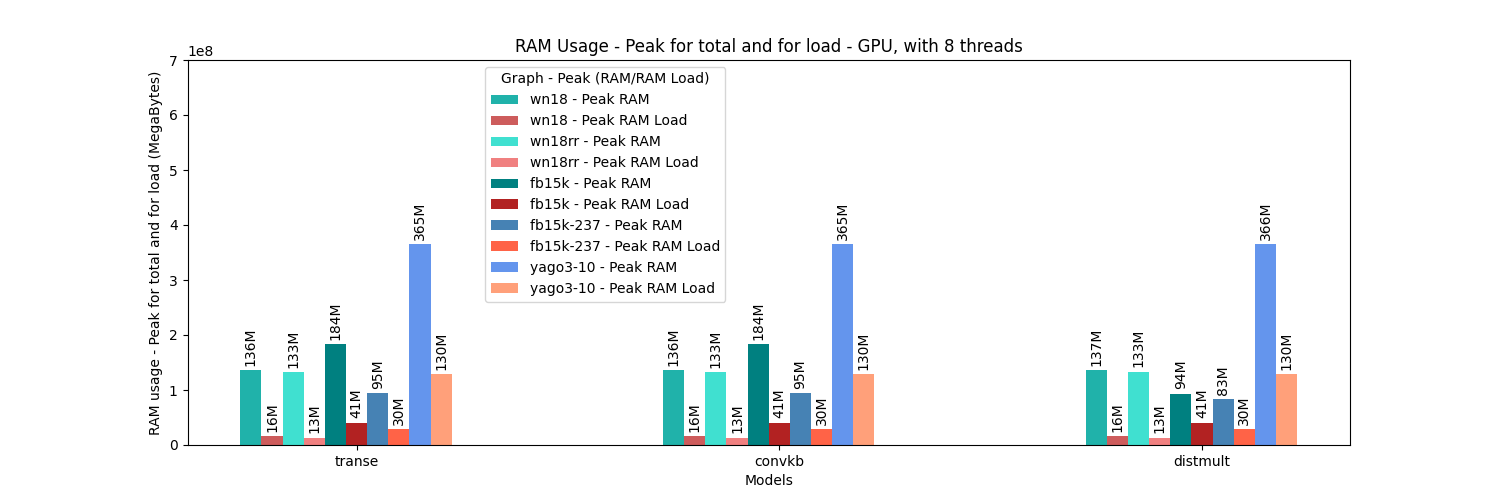,width=3.34in}}
\caption{On the y-axis: RAM usage (in MegaBytes) for total peak and peak of loading graph. On the x-axis: all the tested graphs, considering a number 8 threads, with all the models. Architecture is GPU.}
\label{fig3.ps}
\end{figure}

Considering GPU environment, it can be noticed that the parallelism introduced by this architecture, of different nature with respect to vectorization enabled on CPU, brings to results that are independent of the number of set threads. In this environment, for architectural constraints, the graphs were trained with 2, 4 and 8 threads. This means that, for the settings that TensorFlow enables in the instruction to determine the number of threads \cite{tensorflow}, when the training is performed the number of dedicated threads is not set, as spreading processes along the available resources is what the GPU already does for its nature. See figures in this section, from figure~\ref{fig_example7.ps}, till figure~\ref{fig_example12.ps}. Considering RAM usage, it can be noticed that the RAM peak for loading the graph is always inline with the previous values, with other architectures. On the contrary, the peak RAM for the total process is inferior with respect to the measured peak of RAM with CPU, with a significant delta; this is because GPU has significantly faster and more advanced memory interfaces as it needs to shift around a lot more data than CPU. See figure~\ref{fig2.ps} and figure~\ref{fig3.ps}.

\section{Conclusions}
\label{end}
The task of the project was to show the behavior of selected models in the process of knowledge graph embedding. On the whole, experimental results show the behavior of the models related to the specific graphs, according to the properties and the characteristics of the datasets. Furthermore, CPU architecture without vectorization, has shown that, varying the number of threads, wall clock time always decreases, as it measures the time taken to perform a job. On the contrary, CPU time always increases, as it measures the time in which the CPU is busy. This time can grow, due to the required amount for parallelization between threads. In the environment of CPU with vectorization enabled, the total peak of consumed RAM and the one for loading the graph don’t change; trends in wall clock and CPU train time are respected, and the whole amount of time is inferior because of the parallelism at level of data offered by this type of architecture. At the end, the last tested architecture is GPU environment, in which number of threads has no effects in the measurement of both CPU and wall clock train time, because of the setting of TensorFlow number of threads, and the fact that the GPU already performs the spreading of processes among the available threads. GPU, considering RAM usage, has a total measured peak very inferior to the one measured with both CPU architectures, because GPU has significantly faster and more advanced memory interfaces.

Multithreading is efficient, but it must be noticed that benefit decreases as the number of threads involved in the computation is increased. This happens because saturation in the execution is reached. Besides, comparing the architectures, GPU proves to be the best architecture for the given task, both for execution time and RAM utilization, even if CPU with some vectorized instructions enabled still behaves well. Both GPU and CPU with vectorization enabled, with a smaller number of threads (i.e. 2, 4, 8), achieve better results considering execution time w.r.t. standard CPU with a greater number of threads (i.e. 8, 16, 32), as it was expected. \newline

Considering RAM, it can be noticed that RAM utilization for the loading of the graph never changes considering different architectures. Besides, RAM utilization depends only on the graph, not on the model as it changes according to the graph (i.e. model variation has not influence). Finally, GPU has a total peak of RAM utilization that is inferior to CPU both with and without vectorization enabled. \newline

Obtained results are not representative of the optimal settings for each graph, as the task was not to influence performances based on the capabilities of convergence of the single model, but to analyze performances of different models without influencing too much by defining the optimal settings. Nevertheless, it must be noticed that the optimization realized in every model can converge in a different number of epochs and that the results presented in this document are related on the performances based on the same settings. This point is significant in the moment in which the trained graph is then used to make inference; therefore, reaching a state of convergence is important in that case. Having used the same number of epochs independently of the fact that the trained model would converge allows to show which model is in general faster in the specific settings, but not which one is faster in realizing on the whole the same specific task (i.e. a model can be faster than another in the same setting of 500 epochs, but it could need a greater number of epochs to converge, while the other one reaches convergence). \newline

This project has clearly shown the superiority of GPU architecture in the execution of the given task; furthermore, interesting working ideas have been found, for example considering performances on inference of models, and also, considering only the GPU architecture, new topics of research for future works.


\begin{acknowledgment}
Thanks go to professor Marco Domenico Santambrogio and PhD Guido Walter Di Donato from NECSTLab at Politecnico di Milano for supporting the development of this project. All tested have been remotely executed on servers of Politecnico di Milano. The course of Advanced Computer Architectures, managed by professor M.D. Santambrogio was the theoretical base that brought the interest into this research project, that has been realized under the direct supervision of PhD Guido Walter Di Donato, as part of his research interests and activities.
\end{acknowledgment}

%
% ...




{\footnotesize
\begin{thebibliography}{0}
  \bibitem{ampligraph}\href{https://docs.ampligraph.org/en/1.3.1/}{AmpliGraph} for models TransE, ConvKB and DistMult {\scriptsize \newline$[$\url{https://docs.ampligraph.org/en/1.3.1/}$]$}
  \bibitem{KnowledgeGraphEmbedding1} {Quan Wang , Zhendong Mao , Bin Wang, and Li Guo. 2017. Knowledge Graph Embedding: A Survey of Approaches and Applications}  
  \bibitem{KnowledgeGraphEmbedding2} {Andrea Rossi, Donatella Firmani, Antonio Matinata, Paolo Merialdo, and Denilson Barbosa. 2016. Knowledge Graph Embedding for Link Prediction: A Comparative Analysis}
  \bibitem{KnowledgeGraphEmbedding3} {Hung Nghiep Tran and Atsuhiro Takasu (2019). Analyzing Knowledge Graph Embedding Methods from a Multi-Embedding Interaction Perspective}
  \bibitem{KnowledgeGraphEmbedding4} {Palash Goyal and Emilio Ferrara. 2017. Graph Embedding Techniques, Applications, and Performance: A Survey}
  \bibitem{KnowledgeGraphEmbedding5} {Srijan Kumar, Xikun Zhang, and Jure Leskovec. 2018. Learning Dynamic Embeddings from Temporal Interaction Networks}
  \bibitem{KnowledgeGraphEmbedding6} {Zi Yin and Yuanyuan Shen. 2017. On the Dimensionality of Word Embedding}
  \bibitem{KnowledgeGraphEmbedding7} {Hee-Geun Yoon, Hyun-Je Song, Seong-Bae Park, Se-Young Park. 2016. A Translation-Based Knowledge Graph Embedding Preserving Logical Property of Relations}
  \bibitem{KnowledgeGraphEmbedding8} {Kristina Toutanova and Danqi Chen. 2015. Observed versus latent features for knowledge base and text inference}
  \bibitem{KnowledgeGraphEmbedding9} {Liwei Cai and William Yang Wang. 2018. KBGAN: Adversarial Learning for Knowledge Graph Embeddings}
  \bibitem{adam} {Diederik P. Kingma, Jimmy Ba. 2014. Adam: A Method for Stochastic Optimization}
  \bibitem{yago3} {Mahdisoltani, Biega, and Suchanek. 2015. YAGO3: A Knowledge Base from Multilingual Wikipedias}
  \bibitem{freebase} {Kurt Bollacker, Colin Evans, Praveen Paritosh, Tim Sturge, and Jamie Taylor. 2008. Freebase: A Collaboratively Created Graph Database for Structuring Human Knowledge}
  \bibitem{wikidata} {Denny Vrandečić and Markus Krötzsch. 2014. Wikidata: A Free Collaborative Knowledgebase}
  \bibitem{yago} {F. M. Suchanek, G. Kasneci, and G. Weikum. 2007. Yago: a core of semantic knowledge.}
  \bibitem{wordnet} \href{https://wordnet.princeton.edu/}{WordNet} {\scriptsize \newline$[$\url{https://wordnet.princeton.edu/}$]$}
  \bibitem{transe} {Antoine Bordes, Nicolas Usunier, Alberto Garcia-Duran. 2013. Translating Embeddings for Modeling Multi-relational Data}
  \bibitem{convkb} {Dai Quoc Nguyen, Tu Dinh Nguyen, Dat Quoc Nguyen, Dinh Phung. 2018. A Novel Embedding Model for Knowledge Base Completion Based on Convolutional Neural Network}
  \bibitem{distmult} {Bishan Yang, Wen-tau Yih, Xiaodong He, Jianfeng Gao and Li Deng. 2015. Embedding entities and relations for learning and inference in knowledge bases}
  \bibitem{pandas}\href{https://pandas.pydata.org/}{Pandas}  for generating plots {\scriptsize \newline$[$\url{https://pandas.pydata.org/}$]$}
  \bibitem{matplotlib}\href{https://matplotlib.org/}{Matploblib}  for generating plots {\scriptsize \newline$[$\url{https://matplotlib.org/}$]$}
  \bibitem{tensorflow}\href{https://www.tensorflow.org/}{TensorFlow} {\scriptsize \newline$[$\url{https://www.tensorflow.org/}$]$}
  \bibitem{repo}{GitHub Valeriani Repository} {\scriptsize \newline$[$\url{https://github.com/AngelicaSofia/ACA-project/}$]$}
  \bibitem{intel}{Intel First Reference} {\scriptsize \newline$[$\url{http://www.obpm.org/download/Intro_to_Intel_AVX.pdf}$]$}
  \bibitem{intel2}{Intel Second Reference} {\scriptsize \newline$[$\url{https://software.intel.com/content/www/us/en/develop/articles/maximize-tensorflow-performance-on-cpu-considerations-and-recommendations-for-inference.html}$]$}
  \bibitem{proto}{TensorFlow config.proto} {\scriptsize \newline$[$\url{https://github.com/tensorflow/tensorflow/blob/26b4dfa65d360f2793ad75083c797d57f8661b93/tensorflow/core/protobuf/config.proto#L165}$]$}
\end{thebibliography}}
\end{document}